\documentclass{article}
\usepackage[utf8]{inputenc}

\usepackage{microtype}
\usepackage{graphicx}
\usepackage{subfigure}
\usepackage{booktabs}
\usepackage{url}

\graphicspath{{./figs/}}
\usepackage{natbib}
\usepackage{multirow}

\usepackage[accepted]{icml2019}

\usepackage[dvipsnames]{xcolor}
\usepackage{soul}

\icmltitlerunning{Cognitive Model Priors for Predicting Human Decisions}

\begin{document}

\twocolumn[
\icmltitle{Cognitive Model Priors for Predicting Human Decisions}
\icmlsetsymbol{equal}{*}

\begin{icmlauthorlist}
\icmlauthor{David D. Bourgin}{equal,ucb}
\icmlauthor{Joshua C. Peterson}{equal,princeton}
\icmlauthor{Daniel Reichman}{princeton}
\icmlauthor{Thomas L. Griffiths}{princeton}
\icmlauthor{Stuart J. Russell}{ucb}
\end{icmlauthorlist}

\icmlaffiliation{ucb}{University of California, Berkeley}
\icmlaffiliation{princeton}{Princeton University}

\icmlcorrespondingauthor{David D. Bourgin}{ddbourgin@gmail.com}
\icmlcorrespondingauthor{Joshua C. Peterson}{peterson.c.joshua@gmail.com}

\icmlkeywords{cognitive models, neural networks, data scarcity}

\vskip 0.3in
]

\printAffiliationsAndNotice{\icmlEqualContribution} 
\begin{abstract}
Human decision-making underlies all economic behavior. For the past four decades, human decision-making under uncertainty has continued to be explained by theoretical models based on prospect theory, a framework that was awarded the Nobel Prize in Economic Sciences. However, theoretical models of this kind have developed slowly, and robust, high-precision predictive models of human decisions remain a challenge. While machine learning is a natural candidate for solving these problems, it is currently unclear to what extent it can improve predictions obtained by current theories. We argue that this is mainly due to data scarcity, since noisy human behavior requires massive sample sizes to be accurately captured by off-the-shelf machine learning methods. To solve this problem, what is needed are machine learning models with appropriate inductive biases for capturing human behavior, and larger datasets. We offer two contributions towards this end: first, we construct ``cognitive model priors'' by pretraining neural networks with synthetic data generated by cognitive models (i.e., theoretical models developed by cognitive psychologists). We find that fine-tuning these networks on small datasets of real human decisions results in unprecedented state-of-the-art improvements on two benchmark datasets. Second, we present the first large-scale dataset for human decision-making, containing over 240,000 human judgments across over 13,000 decision problems. This dataset reveals the circumstances where cognitive model priors are useful, and provides a new standard for benchmarking prediction of human decisions under uncertainty.
\end{abstract}

\section{Introduction}
\label{intro}

Which gamble would you rather take: a 50/50 chance of winning/losing \$100, or a 100\% chance of \$0? Although both gambles have equal expected value (payoff), the majority of people systematically prefer the second \cite{kahneman1979prospect}.

Predicting the choices that people make in such situations is of central importance in economics and relevant to understanding consumer behavior. As suggested above, a recurring empirical finding is that people reliably deviate from optimal decision-making under various conditions. Extensive research has sought to understand these deviations and the conditions that elicit them (\citealt{edwards1954theory,baron2000thinking,gilovich2002heuristics}, to name just a few). A prime example is prospect theory, devised by \citet{kahneman1979prospect}, and a key part of Kahneman's receipt of the Nobel Prize in Economic Sciences. Constructing models of choice behavior based on prospect theory and its descendants (e.g., \citealt{tversky1992advances,erev2017a}) remains an active research area in behavioral economics and cognitive psychology. Good predictive models of human choice also arise in AI applications as they can be used to build artificial agents that are better aligned with human preferences \cite{rosenfeld2018predicting,russell2016artificial,chajewska2001learning}.

Despite these efforts, we are still far from accurate and robust models of how people make decisions. Theoretical explanations for deviations from maximizing expected value are often contradictory, making it difficult to come up with a single framework that explains the plethora of empirically observed deviations such as loss aversion, the St. Petersburg paradox, and many others \cite{erev2017a}. This state of affairs has motivated several recent choice prediction competitions \cite{erev2010choice,erev2017a,plonsky2019predicting} in an attempt to achieve better predictive models of human decisions.

One approach towards improving prediction has been to leverage machine learning (ML)~\cite{rosenfeld2018predicting,plonsky2017a,noti2017behavior,plonsky2019predicting}. However, data scarcity has remained the key limiting factor in leveraging ML to predict human decisions. Cognitive models of human decision-making (e.g., prospect theory) are typically developed by behavioral scientists using datasets consisting of hundreds or fewer data points. Such small datasets, coupled with the complex and noisy nature of human behavior, increase the risk of overfitting and limit the applicability of machine learning methods \cite{geman1992neural,yarkoni2017choosing}.

In this work we develop and evaluate a method for enabling machine learning to achieve better predictions of human choice behavior in settings where data are scarce. Our approach is to treat cognitive models as a source of inductive bias to help machine learning methods ``get off the ground,'' since their predictions are closer to human behavior than untrained machine learning models. The resulting ``cognitive model priors'' offer a solution to the problem of prediction under data scarcity by combining the rich knowledge available in cognitive models with the flexibility to easily adapt to new data.

\noindent\textbf{Contributions:}    

Our contributions are as follows:

\begin{itemize}
\itemsep0.18em
    \item 
  We introduce a simple and general methodology for translating cognitive models into inductive biases for machine learning methods. At a high level, our method consists of generating \emph{synthetic datasets} generated from cognitive models that can be used to establish informative priors for ML algorithms. We focus on the case where a neural network is trained on these synthetic datasets and then fine-tuned using a much smaller dataset of real human data.
  \item
  We test this methodology using two new large synthetic datasets (\texttt{synth15} and \texttt{synth18}) that capture current theoretical knowledge about human decision-making. Transferred to ML models, the resulting cognitive model priors provide a new way to predict human decisions that integrates the flexibility of ML with existing psychological theory.
  \item
  Using this new methodology we greatly improve state-of-the-art performance on two recent human choice prediction competition datasets \cite{erev2017a,plonsky2019predicting}. The resulting models operate on raw features alone (not relying on hand-designed, theory-based features), for the first time for a competitive machine learning model of human decisions.
\item 
  We introduce a new benchmark dataset for human choice prediction in machine learning that is an order of magnitude larger than any previous datasets, comprising more than 240,000 human judgments over 13,000 unique decision problems. We show that even in datasets of this magnitude, cognitive model priors reduce prediction error and increase training efficiency.
\end{itemize}

\noindent\textbf{Related Work:}
There has been a growing interest in applying machine learning to problems in behavioral economics \cite{camerer2018artificial,peysakhovich2017using,hartford2016deep} and econometrics \cite{mullainathan2017machine}. Several recent works have leveraged ML to gain new theoretical insights
\cite{peysakhovich2017using,fudenberg2018predicting,kleinberg2017theory}, while other work has focused solely on improving prediction. For example, \citet{hartford2016deep} introduced an inductive bias in the form of a specialized neural network layer, allowing for improved prediction in the face of scarce training data. The particular bias proposed, however, was only applicable to the domain of two-player games.

Most closely related to our work is that of \citet{plonsky2017a}, who applied an array of popular ML algorithms to the problem of predicting human decisions for pairs of gambles (the same gambles we will consider in this paper). In addition to providing baseline performance results for a number of ML algorithms on human decision data, the authors used significantly more data than was present in previous work \citep[e.g.,][]{peysakhovich2017using}. Notably, the authors found that the predictions made using these algorithms were much poorer than a baseline model based on psychological theory when the raw gambles were used as input. Interestingly, however, when model inputs were supplemented by the components of the theoretical model (expressed as features), one of the ML models---a random forest algorithm---showed an improvement over the baseline psychological model. The authors also found that using theory-based predictions as additional inputs improves performance even further \citep{plonsky2017a,plonsky2019predicting}. These approaches differ from ours in that they require a model-to-feature decomposition in order to leverage the theoretical model, which may not be unique, and can require significant effort and expert knowledge.

\section{Decision-Making Under Uncertainty}
\label{decision-making}

Human decision-making under uncertainty has traditionally been studied in the context of sets of gambles with uncertain outcomes. A single gamble is a set of $N$ possible outcomes $x_i$ and their probabilities $p_i$. Given a set of $M$ gambles, the decision-maker must choose the single most appealing option. In the remainder of the paper, we will be concerned with learning a model that can predict the probability that a human decision-maker will choose each gamble. We want to infer a probability as opposed to a binary selection in order to capture variability both within a single person's choice behavior and across different people. 

\subsection{Cognitive Models of Decision-Making}
\label{cog-models}

Given a utility function $u(\cdot)$ that reflects the value of each outcome to a decision-maker, the rational solution to the choice problem is to simply choose the gamble that maximizes expected utility (EU): $\sum_{i=1}^N{p_iu(x_i)}$ \cite{morgenstern1944theory}. Interestingly, humans do not appear to adhere to this strategy, even when allowing for a range of utility functions. In fact, cognitive models of human decision-making were originally designed in order to capture four \textit{deviations} from EU, with the most influential model being prospect theory \citep{kahneman1979prospect}. This model asserts that people assess a quantity that takes a similar form to expected utility, $V=\sum_{i=1}^N{\pi({p_i)}v(x_i)}$, but where $\pi(\cdot)$ is a weighting function that nonlinearly transforms outcome probabilities and $v(\cdot)$ is a subjective assessment that can be influenced by factors such as whether the outcome is perceived as a gain or a loss.

Since prospect theory was first introduced, the list of discovered human deviations from expected utility has grown significantly, making it increasingly difficult to capture them all within a single unifying model \cite{erev2017a}. One of the most successful recent attempts to do so is the Best Estimate and Sampling Tools (BEAST) model, which eschews subjective weighing functions in favor of a complex process of mental sampling. BEAST represents the overall value of each prospect as the sum of the best estimate of its expected value and that of a set of sampling tools that correspond to four behavioral tendencies. As a result, gamble A will be strictly preferred to gamble B if and only if:

\begin{equation} 
    \left[BEV_{\mathrm{A}} - BEV_{\mathrm{B}}\right] + \\ 
    \left[ST_{\mathrm{A}} - ST_{\mathrm{B}}\right] + e > 0,
\end{equation} 

where $BEV_{\mathrm{A}} - BEV_{\mathrm{B}}$ is the advantage of gamble A over gamble B based on their expected values, $ST_{\mathrm{A}} - ST_{\mathrm{B}}$ is the advantage based on alternative sampling tools, and $e$ is a normal error term.

Sampling tools include both biased and unbiased sample-based estimators designed to explain four psychological biases: (1) the tendency to assume the worst outcome of each gamble, (2) the tendency to weight all outcomes as equally likely, (3) sensitivity to the sign of the reward, and (4) the tendency to select the gamble that minimizes the probability of immediate regret.\footnote{For further details, see \citet{erev2017a}, and source code at:\\ \url{cpc-18.com/baseline-models-and-source-code/}} In the remainder of the paper, we use BEAST as a proxy for current theoretical progress as it both explains a large number of behavioral phenomenon and was built explicitly to address concerns about the applicability of theoretical models to robust prediction.

\subsection{Choice Prediction Competitions}
\label{cpc}

Two important resources for evaluating predictive models of human decision-making are the 2015 and 2018 Choice Prediction Competition datasets \citep[CPC15 and CPC18, respectively; ][]{erev2017a,plonsky2019predicting}.
These competitions offer two benefits. First, they encompass a large space of sets of gambles. Whereas behavioral experiments tend to study only one phenomenon at a time, the CPC datasets were explicitly designed to include sets of gambles that elicit all known deviations from EU in addition to other gambles sampled from a much larger problem space. Second, although we will later argue that such datasets are still too small to fully train and evaluate predictive models, they are currently the largest of their kind. CPC15 contains 90 choice problems (30 test problems) for five repeated trials for a total of 450 datapoints (150 test datapoints), and CPC18 (a superset of CPC15) contains 210 choice problems (90 test problems) for five repeated trials for a total of 1,050 datapoints (450 test datapoints).

Problems in the CPC datasets required people to choose between a pair of gambles, A and B. Each gamble consisted of a collection of rewards and their associated outcome probabilities. In CPC15, gamble A was constrained to only have only two outcomes (similar to the example given in section \ref{decision-making}). Gamble B yielded a fixed reward with probability $1-p_L$ and the outcome of a lottery (i.e., the outcome of another explicitly described gamble) otherwise. That is, by convention in the competition problems, a lottery is defined as the outcome of a chosen gamble (occurring with probability $p_L$) that can also yield one of multiple monetary outcomes, each with some probability. Gamble B's lottery varied by problem and was parameterized using a range of options, including the number of outcomes and the shape of the payoff distribution. In CPC18, the only difference was that some of the problems allowed gamble A to take on the more complex lottery structure of gamble B. 

For each problem, human participants made sequential binary choices between gamble A and B for five blocks of five trials each. The first block of each problem was assigned to a no-feedback condition, where subjects were shown only the reward they received from the gamble they selected. In contrast, during the remaining four blocks subjects were shown the reward they obtained from their selection as well as the reward they could have obtained had they selected the alternative gamble. Finally, gambles for problems assigned to an ``ambiguous'' condition had their outcome probabilities hidden from participants. Further details on the design and format of the gamble reward distributions can be found in \citet{erev2017a} and \citet{plonsky2019predicting}. Prediction of the aggregated human selection frequencies (proportions between 0 and 1) is made given a 12-dimensional vector of the parameters of the two gambles and the block number. As this information is all displayed to the participant in the course of their selection, we refer to these as ``raw'' problem features. Prediction accuracy is measured using mean squared error (MSE).

To summarize, both CPC15 and CPC18 contain choice problems separated into training and test sets. The task is to build or train a model to predict the probability (i.e., the proportion of human participants) that selected gamble A for each problem based on the training set, and evaluate on the test set using MSE.

\subsection{Data Scarcity in Behavioral Sciences}
\label{scarcity}

Scientific studies with human participants often yield small datasets. One reason for this is resource limitations: compensating participants is costly, and large-scale data collection can be limited by the ability to recruit participants. A second reason is that small datasets are a byproduct of the theory-building process: because it is rare to know all the relevant variables ahead of time, researchers must isolate, manipulate, and analyze manageable sets of independent variables over many studies rather than collect a single large dataset.

The high variability of human behavior also makes prediction challenging. Indeed, many important types of human behavior (e.g., decision making) differ significantly across individuals and require large sample sizes to accurately assess. For example, the number of problems in CPC15 and CPC18 for which human data could be feasibly collected was limited because obtaining stable estimates of aggregate behavior required a large number of human participants per choice problem.

Small datasets and high variability may help to explain why machine learning approaches have garnered only marginal improvements in predicting human decisions.
Given these challenges, it is worth considering possible sources of human-relevant inductive biases that might help alleviate the symptoms of data scarcity.

\section{Cognitive Model Priors}

Scientific theory-building is principally concerned with \textit{understanding} and \textit{explanation}, whereas machine learning is geared more towards \textit{prediction}, often employing complex and opaque models. Given enough data, and assuming the domain is of tractable complexity, a machine learning model might rediscover some of the components of an established scientific theory, along with a number of new, potentially complicated and nuanced improvements that may be hard to immediately explain, but are responsible for robust and accurate forecasting.

When data is scarce, however, theoretical models (which tend to be relatively parsimonious compared to models from ML) offer a valuable source of bias, resulting in inferior accuracy but potentially superior generalization \cite{yarkoni2017choosing}. Ideally, we would like to find a way to leverage the hard-won generalizability of psychological theories while making use of the flexibility of machine learning methods to develop more powerful predictive models.

Towards this goal, we propose building a bridge between the two modeling domains via the common currency of data. Assuming a cognitive model can be expressed as some function $f(\cdot)$, where inputs are experimental task descriptions or stimuli and outputs are human choices, preferences, or behaviors, we can convert the insights contained in the scientific model into a synthetic dataset and new model as follows:

\begin{enumerate}
  \item Evaluate a large range of inputs, including those without accompanying human targets, to generate input-target pairs ($x_i$, $f(x_i)$) for training.
  \item Train a machine learning model to approximate $f(\cdot)$ via ($x_i$,$f(x_i)$) as opposed to approximating human decision functions directly ($x_i$ and human targets $h_i$).
  \item Fine-tune the resulting model on small, real human datasets ($x_i$,$h_i$) in order to learn fine-grained improvements to the scientific model.
\end{enumerate}

By instantiating theoretical knowledge in a synthetic dataset, we provide a domain-general way to construct a {\em cognitive model prior} for ML algorithms. This prior can be explicit (conjugate priors for Bayesian models can be interpreted as encoding the sufficient statistics of previous experience \citep{box2011bayesian}) or implicit (if used for pretraining neural networks, those networks will be regularized towards pretrained weights rather than an arbitrary set of initial weights).

\subsection{Converting Choice Prediction Models to Data}

We sampled approximately 100k new problems from the space of possible CPC15 problems and 85k problems from the space of possible CPC18 problems using the procedures from \citet{plonsky2017a} (Appendix D) and \citet{plonsky2018a} (Appendix D). These datasets of sampled problems are at least two orders of magnitude larger than those generated previously using these procedures meant for human experiments. We found these dataset sizes to be both feasible to generate and sufficient to allow off-the-shelf ML methods to learn good approximations to cognitive models. For each new dataset we ensured that no problem occurred more than once or overlapped with the existing CPC15 or CPC18 problems, and removed ``degenerate'' problems in which either both gambles had the same distribution and payoffs or at least one gamble had no variance, but the rewards for the two gambles were correlated. We denote these new collections of problems as the \texttt{synth15} and \texttt{synth18} datasets, respectively.

To create training targets for our cognitive model prior, we used two publicly available versions of the BEAST model (section \ref{cog-models}. These include the original BEAST model proposed before CPC15, which we denote as \texttt{BEAST15}, and the subsequent ``Subjective Dominance'' variant introduced before CPC18, which we denote as \texttt{BEAST18}. These two models were used to predict gamble selection frequencies for each problem in the \texttt{synth15} and \texttt{synth18} datasets respectively. These predictions could then be used as training targets for our machine learning model in order to transfer insights in the theoretical models into a form that can be easily modified given new data to learn from.

\subsection{Model Setup and Fine-Tuning}
\label{model-setup}

We opted to use neural networks to transfer knowledge from BEAST because they can be easily fine-tuned to real human data after initial training and provide a number of different options for regularization. While sequential learning of this sort risks catastrophic forgetting \cite{french1999catastrophic}, we found that fine-tuning with a small learning rate (1e-6) was sufficient to make use of prior knowledge about BEAST internalized in the network's weights. This allows for a straightforward integration of cognitive model priors into the standard neural network training paradigm.

\noindent \textbf{Neural Network Architecture.} We grid-searched approximately $20{,}000$ hyperparameter settings and found the best multilayer perceptron (MLP) overall for estimating both variants of BEAST (as well as the other datasets/tasks in this paper) had three layers with $200$, $275$, and $100$ units respectively, SReLU activation functions, layer-wise dropout rates of $0.15$, and an RMSProp optimizer with a $0.001$ learning rate. The output layer was one-dimensional with a sigmoid activation function to match the range of the human targets. We also obtained lower error, less overfitting, and more stable fine-tuning using the SET algorithm \cite{mocanu2018scalable}, which consists of (1) initializing the network as an Erdős--Rényi random sparse graph and (2) evolving this graph at the end of each epoch via both pruning of small weights and the addition of new random connections. While we don't view this architectural choice as essential for the application of cognitive model priors, this form of model compression may be additionally useful when data is scarce. 

\subsection{Results}
\label{cpc-results}

\textbf{CPC15:} The results of our analysis along with the performance of several baseline and competing models (drawing on \citealt{plonsky2017a}) are given in Table \ref{cpc-table}. For example, a wide range of common machine learning algorithms utterly fail to obtain reasonable MSE scores given the raw gamble parameters shown to human participants, likely due to a lack of data. For CPC15, \texttt{BEAST15} was the competition's theoretical baseline model, and beats all unaided ML methods. The competition organizers also note that no other theory-based models from psychology or behavioral economics, such as prospect theory models, reached the leaderboard's top-10.\footnote{\url{http://departments.agri.huji.ac.il/economics/teachers/ert_eyal/compres.htm}} The CPC15 winner was a small augmentation of \texttt{BEAST15} with no machine learning component. The table section \textit{ML + Feature Engineering} shows results for the method employed in \citet{plonsky2017a}, wherein intermediate features derived from \texttt{BEAST15} were used as input to ML algorithms as opposed to raw features alone (i.e., parameters of the gambles). Notably, these models also do worse than the self-contained \texttt{BEAST15} model with the exception of the random forest model, which lowered test set MSE of the CPC15 winner by 0.0001. More successful was an ensemble that included the random forest model with the \texttt{BEAST15} prediction as an additional feature, obtaining an MSE of 0.007. In contrast to all of these baselines, our own method of fine-tuning a neural translation of \texttt{BEAST15} by way of \texttt{synth15} obtains a large reduction in MSE to 0.0053 (24\% decrease).
\vspace{-3mm}
\begin{table}[!ht]
\centering
\caption{Performance (MSE) for CPC15 and CPC18 benchmarks.}
\label{cpc-table}
\begin{tabular}{llc}
\bottomrule \\ [-2.0ex]
& Model & MSE$\times$100 \\ [-0.3ex]
\bottomrule \\ [-1.5ex]
\multirow{15}{*}{\rotatebox[origin=c]{90}{\textbf{----------------- \,CPC 2015\, -----------------}}} & \textit{ML + Raw Data} & \\
& \,\,\,\,\,\,\,\,MLP & 7.39 \\
& \,\,\,\,\,\,\,\,$k$-Nearest Neighbors & 7.15 \\
& \,\,\,\,\,\,\,\,Kernel SVM & 5.52 \\
& \,\,\,\,\,\,\,\,Random Forest & 6.13 \\
& \textit{Theoretical Models} & \\
& \,\,\,\,\,\,\,\,\texttt{BEAST15} & 0.99 \\
& \,\,\,\,\,\,\,\,CPC 2015 Winner & 0.88 \\
& \textit{ML + Feature Engineering} & \\
& \,\,\,\,\,\,\,\,MLP & 1.81 \\
& \,\,\,\,\,\,\,\,$k$-Nearest Neighbors & 1.62 \\
& \,\,\,\,\,\,\,\,Kernel SVM & 1.01 \\
& \,\,\,\,\,\,\,\,Random Forest & 0.87 \\
& \,\,\,\,\,\,\,\,Ensemble & 0.70 \\
& \textbf{MLP + Cognitive Prior (ours)} & \textbf{0.53} \\ [0.6ex]
\midrule \\ [-2.0ex]
\multirow{6}{*}{\rotatebox[origin=c]{90}{\textbf{--- \,CPC 2018\, ---}}} & \textit{Theoretical Models} & \\
& \,\,\,\,\,\,\,\,\texttt{BEAST18}  & 0.70 \\
& \textit{ML + Feature Engineering} & \\
& \,\,\,\,\,\,\,\,Random Forest     & 0.68 \\
& \,\,\,\,\,\,\,\,CPC 2018 Winner   & 0.57 \\
& \textbf{MLP + Cognitive Prior (ours)} & \textbf{0.48} \\ [0.5ex]
\bottomrule
\end{tabular}
\end{table}
\vspace{-2mm}

\textbf{CPC18:} While the same set of baselines are not available for CPC18, the most crucial comparisons are given in the second half of Table \ref{cpc-table}. First, \texttt{BEAST18}, a modification of \texttt{BEAST15}, obtains a fairly low MSE score of 0.007. Since CPC15 is a subset of CPC18, this can be considered an improvement. Second, \citet{plonsky2018a} released a random forest baseline that makes use of theory-inspired input features, given it was the only successful application of ML in CPC15. Like CPC15, a small decrease in MSE of 0.0002 is obtained. The winner of the CPC18 competition---a submission by the authors of the current paper---was a gradient-boosted decision tree regressor and obtained a notable decrease in MSE for the first time using ML methods. Part of this success is likely due to the larger CPC18 training set. Finally, the further improvement we provide in this paper is a \texttt{synth18}-trained MLP that was fine-tuned on the CPC18 training set that produced an MSE of 0.0048, a decrease in error of 0.0022 (16\%) over the competition's theoretical baseline. The resulting model (as well as our MLP for CPC15) take only raw gamble parameters as input when fully trained.

\section{A New Human Decision Dataset}
\label{choices13k}

\begin{figure}[!ht]
    \centering
    \includegraphics[trim = 20mm 0mm 20mm 20mm, clip,origin=c,width=1.0\linewidth]{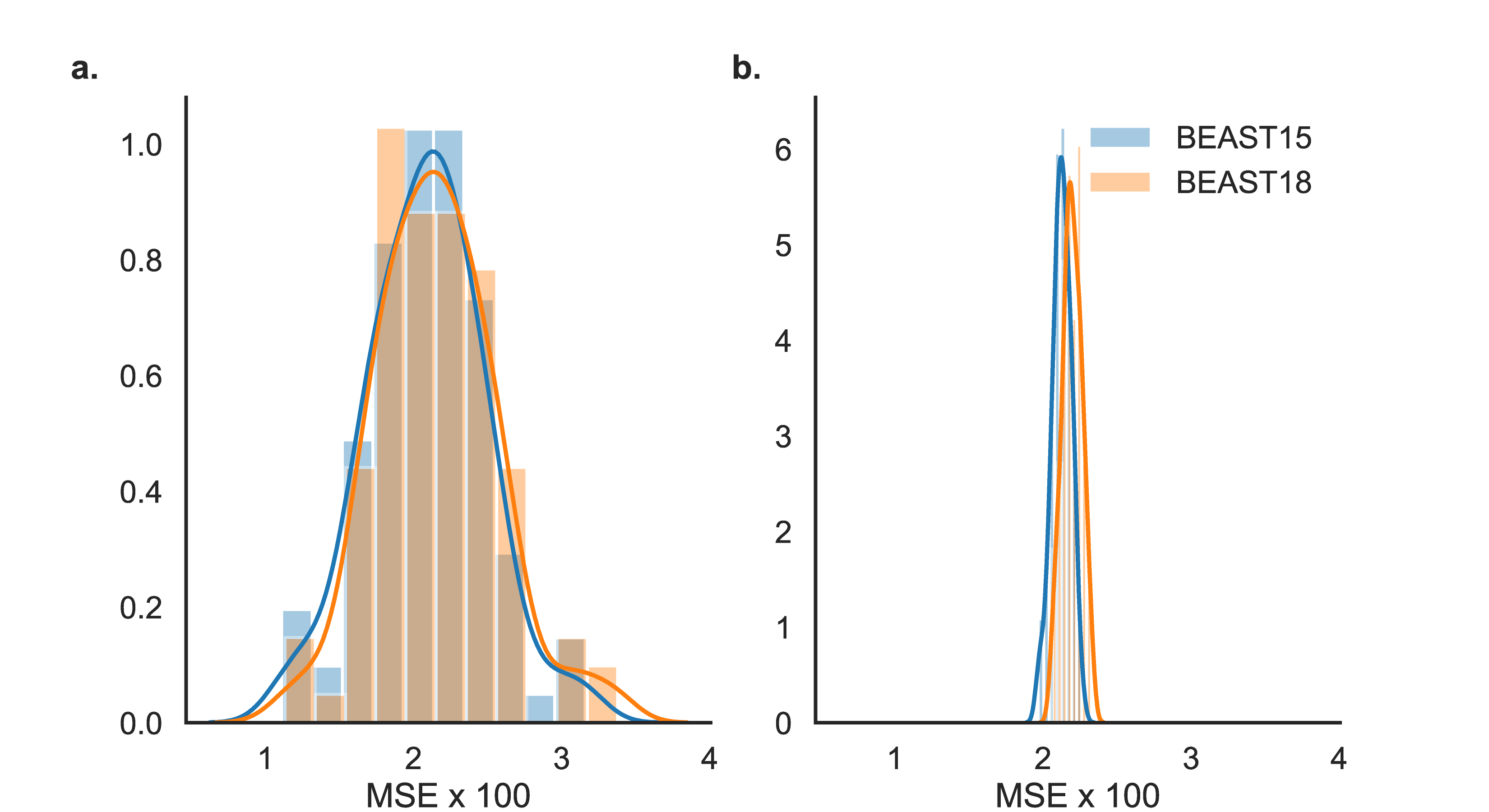}
    \caption{\textbf{Left:} The distribution of BEAST MSEs over 100 bootstrapped samples of size equal to the CPC 2018 training dataset (210 problems). \textbf{Right:} The distribution of BEAST MSEs on 50\% of the entire \texttt{choices13k} dataset (approx. 6,500 problems).}
    \label{fig:bootstrap}
\end{figure}

In many ML competitions, the published test sets often begin to function as unofficial validation sets, eventually leading to validation overfitting \cite{recht2018cifar}. Further, CPC test sets are much smaller than most benchmark datasets in machine learning, making for a poor overall test of generalization. Thus, even though cognitive model priors may make it possible to train on small datasets, a larger dataset is still necessary to establish their value with confidence. In the following section we introduce \texttt{choices13k}, a new large-scale dataset of human risky-choice behavior, and use it to evaluate our cognitive model priors. Before doing so, however, we preview a crucial initial result on this dataset to underscore the risk of using small test sets. In Figure \ref{fig:bootstrap} we drew one-hundred bootstrap samples from our larger dataset to simulate variability in small samples of validation problems. When drawing samples the size of CPC18, the variability in fit by \texttt{BEAST18} is large, and stabilizes considerably when taking samples of up to approximately 6,500 problems. This motivates the need to obtain a larger validation set for assessing the utility of cognitive model priors, and also provides an opportunity to assess whether such priors have a benefit even when data is in abundance.

\subsection{Data Collection for \texttt{choices13k}}

We collected a dataset more than ten times the size of CPC18 using Amazon Mechanical Turk. In total, the new dataset contained 242{,}879 human judgments on 13{,}006 gamble selection problems, making it the largest public dataset of human risky choice behavior to date. The gamble selection problems were sampled uniformly at random from the \texttt{synth15} dataset. The presentation format for the gambling problems was inspired by the framework used in the 2015 and 2018 Choice Prediction Competitions \citep{erev2017a,plonsky2019predicting}. The only exceptions in our datasets is that the \texttt{block} parameter in CPC15 and CPC18 was reduced from five (ie., subjects completed five blocks of five trials for each problem) to two, and that feedback and no-feedback blocks were not required to be presented sequentially. We found that this alteration did not significantly affect overall predictive accuracy in our models, and allowed us to more than halve the number of trials necessary for each problem.

\begin{figure}[!t]
    \centering
    \includegraphics[trim = 0mm 0mm 0mm 0mm, clip,origin=c,width=1.0\linewidth]{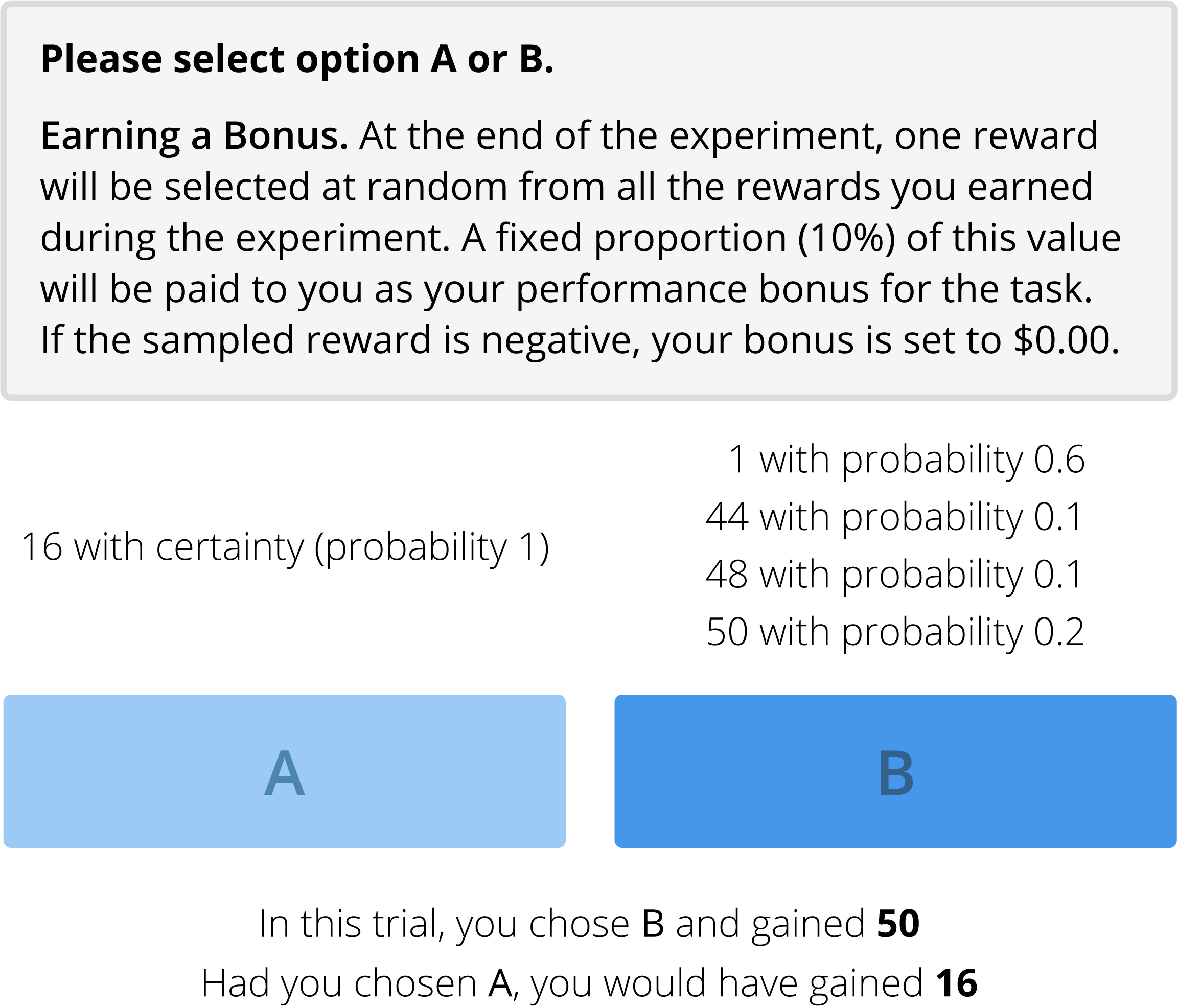}
    \caption{Experiment interface for collecting human gamble selections, modeled after the interface in \citet{erev2017a}.}
    \label{fig:interface}
    \vspace{-1mm}
\end{figure}

As in the CPC datasets, each problem required participants to choose between two gambles, after which a reward was sampled from the selected gamble's payoff distribution and added to the participant's cumulative reward (Figure \ref{fig:interface}). Participants that selected gambles on the same side of the screen over more than 80\% of the trials were excluded. In all, the final dataset consisted of gamble selection frequencies for an average of 16 participants per problem. Each participant completed 16 problems in the feedback condition and four problems in the no feedback condition and was paid \$0.75 plus a bonus of 10\% of their winnings from a randomly selected problem.

\begin{figure*}[!ht]
    \centering
    \includegraphics[trim = 18mm 0mm 24mm 12mm, clip,origin=c,width=1.0\linewidth]{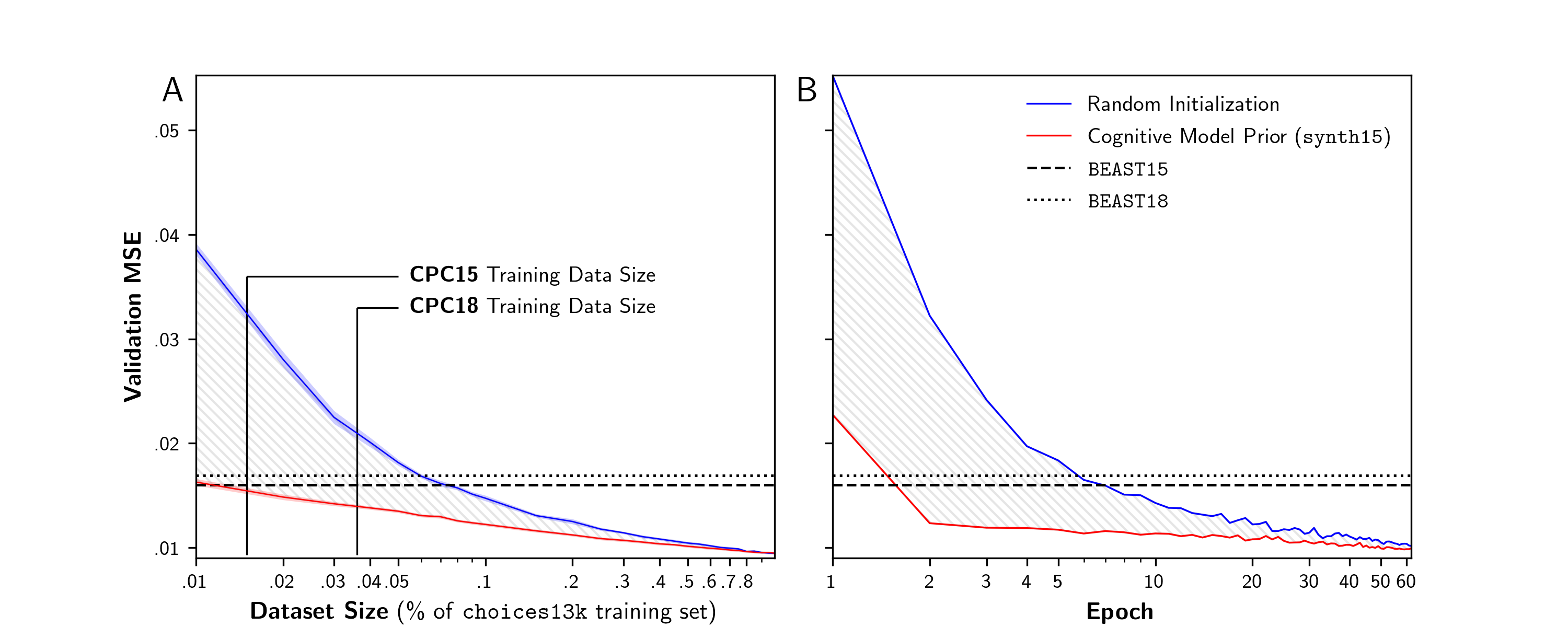}
    \caption{\textbf{A.} Validation MSE (20\% of full \texttt{choices13k} human dataset) as a function of training on proportions of the training set (80\% of full \texttt{choices13k} human dataset), for sparse MLPs trained from a random initialization (blue), and with a cognitive model prior (red). The prior allows for comparable MSE with significantly less data. \textbf{B.} Validation MSE (20\% of full \texttt{choices13k} human dataset) as a function of training epoch. Using a prior allows for faster training. Both x-axes are shown in log-scale.}
    \label{fig:prior_plus_bigdata}
\end{figure*}

\subsection{Assessing the Value of Cognitive Model Priors}
The expanded \texttt{choices13k} dataset allows us to quantify the predictive power of psychological theory, and to assess the relationship between data scarcity and the influence of our cognitive model prior. To this end, we varied the amount of training data available to neural networks that were either randomly initialized or pretrained on the \texttt{synth15} cognitive model prior. The training sets we varied were a proportion (from 0.01 to 1.0) of our full \texttt{choices13k} training set (80\% of the overall dataset). The remaining 20\% of \texttt{choices13k} was used as a constant validation set and the size did not vary. We repeated this process ten times.

Figure \ref{fig:prior_plus_bigdata}A plots validation MSE for a sparse MLP trained from a random initialization (blue) and one pretrained on \texttt{synth15} (i.e., using a cognitive model prior). Standard error contours of the average MSE scores, computed over ten train/validation splits, are shown in lighter colors. Note that models were re-initialized for each split and so variation due to initialization is a subset of the reported error contour regions. The MLP with a cognitive model prior begins with a significantly lower MSE in comparison to the randomly initialized model (0.016 versus 0.039), and continues to exhibit better error scores all the way up to 100\% of the \texttt{choices13k} training set (0.00945 versus 0.00947). The advantage at 25\% of the training data is a decrease by 0.0009, the advantage at 50\% is a decrease by 0.0003, and the advantage at 75\% is a decrease by 0.0002. It takes just 4\% of our real human data (an amount still larger than CPC18) to cross over from an MSE above the best obtained by all other types of machine learning algorithms we tried (see section 4.3), to one below it, a reduction from 0.020 to 0.0138. These results suggests that cognitive model priors may be useful in any setting where collecting human decisions is costly, and further supports the pattern of results in section \ref{cpc-results}.

Beyond overall error, Figure \ref{fig:prior_plus_bigdata}B shows validation MSE per epoch given the full \texttt{choices13k} training dataset. Even though MSE for the two types of models largely converges (with a lingering numerical advantage in final MSE when using the prior), the MSE for the MLP with a cognitive model prior starts lower, decreases much faster, and requires fewer overall epochs to converge. This suggests one advantage of cognitive model priors that we did not anticipate: faster training. While datasets of this size do not require models that take as long to train as, e.g., large deep neural networks, we found this property extremely useful in allowing for larger hyperparameter grid searches.

\subsection{A Benchmark for Predicting Human Decisions}
\label{sec:choices13k_benchmark}
Given the size of our dataset relative to previous ones, we propose it as the new benchmark for validating the precision and generalization of both theoretical and machine learning models. To this end, we provide a set of starting baseline scores as a challenge to ML practitioners (see Table \ref{choices13k-benchmarks}). Notably, while the authors of \citet{plonsky2017a} achieved the best results on their dataset using random forest models, we found neural networks were much more effective for our larger dataset. Our most successful model, the result of a grid-search over 20,000 hyperparameter combinations, was a sparse MLP as described above, containing less than 10,000 parameters (down from around 100,000 for the best full dense MLP). By the end of training (using all \texttt{choices13k} data), the \texttt{synth15} cognitive model prior only added a slight numerical advantage in MSE. The advantage gained by enforcing sparsity likely indicates that our dataset, while much larger than any other of its kind, may still benefit from additional methods for accommodating data scarcity. Given the relatively generic (i.e., off-the-shelf) models used here, we expect there is much room for improvement in MSE on our dataset, particularly with specialized neural architectures combined with new ways of exploiting our cognitive model priors.
\vspace{-4mm}
\begin{table}[!ht]
\centering
\caption{Baseline validation MSE scores on the \texttt{choices13k} benchmark, averaged over ten 80/20 splits.}
\label{choices13k-benchmarks}
\begin{tabular}{lr}
\toprule
                 Model & MSE$\times$100 \\
\midrule
    Linear Regression                    & 4.02 \\
    $k$-Nearest Neighbors                & 2.27 \\
    Kernel SVM                           & 2.16 \\
    Random Forest                        & 1.41 \\
    MLP                                  & 1.03 \\
    Sparse MLP \cite{mocanu2018scalable} & \textbf{0.91} \\
\bottomrule
\end{tabular}
\end{table}
\vspace{-3mm}

\section{Discussion}
\label{discussion}

In this paper we have provided a method for addressing the data scarcity problem that limits the application of machine learning methods to small behavioral datasets by using cognitive models as a source of informative priors. We have outlined a novel approach for doing just that: train a neural network on a large synthetic dataset generated from a cognitive model and use the resulting weights of the model as a \emph{cognitive model prior} that is then fine-tuned on additional human data. Following this approach we were able to efficiently outperform previously proposed cognitive and machine learning methods in predicting human decisions. We believe that the ideas suggested here may prove useful to other behavioral domains such as social cognition, marketing, and cognitive neuroscience.

The weights of the pretrained networks that encode the cognitive model priors could potentially be useful even in broader contexts if they encode a general value function for gambles. For example, one could imagine adding a new input layer with $M>2$ gambles to choose from, which could be trained in isolation before fine-tuning the full network.

The utility of cognitive model priors for human choice prediction raises a number of questions. For example, how ``complex" must the cognitive model prior be in order to produce a marked improvement in prediction error? Does taking EU or prospect theory as the source of a prior suffice to improve upon state-of-the-art prediction? Beyond theoretical interest, this question might have practical applications as it may indicate how much effort to invest in theory-driven models before they could be used as a source of priors. Further, our application of the prior was in the form of weight initialization (after pretraining), but there are many other possibilities, such as an auxiliary loss term for BEAST targets during fine-tuning, or an L2 weight prior centered on the initial weights. Both of these modifications may improve generalization.

Finally, although our focus here has been on prediction, there are many other important ways in which models of human decision-making are used (e.g., providing explanations, interpreting behavior, inferring causality, etc). We do not claim that models of human behavior should be evaluated based solely on their predictive accuracy. Indeed, the current approach trades some of the interpretability of the theoretical models in \citet{plonsky2019predicting} for the ease of integrating that theoretical knowledge into a more powerful predictive modeling framework. However, we do believe that better predictive models, even those driven by data rather than theory, can improve our understanding of human behavior: if machine learning methods can reach or exceed the predictive power of psychological theories, they may be able to point theoreticians toward regularities in behavior that have not yet been detected by human scientists.

\section{Conclusion}
In the current paper we introduced an approach to incorporating theory-based inductive biases from a cognitive model into machine learning algorithms. We demonstrated this approach with two new synthetic datasets: \texttt{synth15} and \texttt{synth18}, and showed that that when data is in short supply, the resulting cognitive model priors allowed our network to achieve significantly better generalization performance with fewer training iterations than equivalent models with no such priors. We also found that integrating these cognitive model priors into a generic neural network model  improved state-of-the-art predictive performance on two public benchmarks, CPC15 and CPC18, without relying on hand-tuned features. Finally, we introduced \texttt{choices13k}, the largest public dataset of human risky choice behavior to date. It is our hope that these three new datasets will support further interaction between the machine learning community and the behavioral sciences, leading to better predictive models of human decision-making.

\section*{Acknowledgments}
This work was supported by grants from the Future of Life Institute, the Open Philanthropy Foundation, DARPA (cooperative agreement D17AC00004), and grant number 1718550 from the National Science Foundation.

\bibliography{references}
\bibliographystyle{icml2019}

\end{document}